# SEMI-AUTOMATED EXTRACTION OF RESEARCH TOPICS AND TRENDS FROM NCI FUNDING IN RADIOLOGICAL SCIENCES FROM 2000-2020

Mark Nguyen[1]*, Peter Beidler[1]*, Joseph Tsai[2], August Anderson[2], Daniel Chen, Paul Kinahan[3,2], John Kang[2]

*these authors contributed equally

[1]University of Washington School of Medicine

[2]University of Washington Dept. of Radiation Oncology

[3]University of Washington Dept. of Radiology

Corresponding author: John Kang, johnkan1@alumni.cmu.edu

**Brief abstract**

Investigators, funders, and the public desire knowledge on topics and trends in publicly funded research but current efforts in manual categorization are limited in scale and understanding. We developed a semi-automated approach to extract and name research topics, and applied this to $1.9B of NCI funding over 21 years in the radiological sciences to determine micro- and macro-scale research topics and funding trends. Our method relies on sequential clustering of existing biomedical-based word embeddings, naming using subject matter experts, and visualization to discover trends at a macroscopic scale above individual topics. We present results using 15 and 60 cluster topics, where we found that 2D projection of grant embeddings reveals two dominant axes: physics-biology and therapeutic-diagnostic. For our dataset, we found that funding for therapeutics- and physics-based research have outpaced diagnostics- and biology-based research, respectively. We hope these results may (1) give insight to funders on the appropriateness of their funding allocation, (2) assist investigators in contextualizing their work and explore neighboring research domains, and (3) allow the public to review where their tax dollars are being allocated.

**Full abstract:**

PURPOSE:

Investigators, funding organizations, and the public desire knowledge on topics and trends in publicly funded research but current efforts for manual categorization have been limited in breadth and depth of understanding. We present a semi-automated analysis of 21 years of R-type NCI grants to departments of radiation oncology and radiology using machine learning (ML) and natural language processing (NLP).

METHODS:

We selected all non-education R-type NCI grants from 2000 to 2020 awarded to departments of radiation oncology/radiology with affiliated schools of medicine. We used pre-trained word embedding vectors to represent each grant abstract. A sequential clustering algorithm assigned each grant to one of 60 clusters representing research topics or domains; we repeated the same workflow for 15 clusters for comparison. Each cluster was then manually named using the top words and closest documents to each cluster centroid. Interpretability of document embeddings was evaluated by projecting them onto 2 dimensions. Changes in clusters over time were used to examine temporal funding trends.

RESULTS:

We included 5,874 non-education R-type grants totaling 1.9 billion dollars of NCI funding over 21 years. Human-model agreement was similar to human interrater agreement. 2-dimensional projections of grant clusters showed two dominant axes: physics-biology and therapeutic-diagnostic. Therapeutic and physics clusters have grown faster over time than diagnostic and biology clusters. The average topic grew at a rate of $46,647 per year over 21 years. The three topics with largest funding increase were *Imaging biomarkers for diagnosis and treatment response*, *Informatics software for treatment decision support*, and *Radiopharmaceuticals*, which all had a mean annual growth >$218,000 per year. The three topics with largest funding decrease were *Cellular stress response*, *Advanced imaging hardware technology* and *Improving performance of breast cancer computer aided detection (CAD)*, which all had a mean decrease of >$110,000 per year.

CONCLUSION:

We developed a semi-automated NLP approach to analyze research topics and funding trends. We applied this approach to NCI funding in the radiological sciences to extract both domains of research being funded and temporal trends. These results may (1) give insight to funders on the appropriateness of their funding allocation, (2) assist investigators in contextualizing their work and explore neighboring research domains, and (3) allow the public to review where their tax dollars are being allocated.

## Introduction

In fiscal year (FY) 2020, the National Institutes of Health (NIH) granted over $36B, of which $6.38B went to awards by the National Cancer Institute (NCI)[1], which awarded $508M of NCI funding to departments of diagnostic radiology and radiation oncology[2,3]. Despite these large sums, it is not entirely clear what areas of study are being funded or how funding is distributed across research topics at a granular level. The NIH has developed tools to help categorize grants for congressional oversight[9]. The NIH Spending Category field was created in 2008 to abide by congressional requests for transparency, though it is very broad (e.g., "lung cancer", "genetics"). Since 2008, the project terms field consists of keywords mined by an automated tool though the nontrivial issue of extracting research topics or directions remains. The NCI publishes an annual Budget Fact Book (last in for FY2020[1]) which includes summary statistics and breaks down research funding by disease site.

Prior investigation into research directions in radiation oncology has relied primarily on surveys and manual curation of grants[4–7], both with regards to identifying topics and assigning grants to these topics. These efforts have categorized NIH funding in radiation oncology during FY 2013 for 197 grants into 3 categories[4], during FY 2010-2012 into 17 biology sub-categories[5], and during FY 2014-2016 for 182 grants 5 topic categories[6] (**Supplementary Text S1**). Collectively, these three studies have iteratively categorized radiation oncology research grants up to 3 years at a time. Due to reliance on expert manual review, these studies are limited to small samples of NIH awards. Each study sorted grants into different categories and used different inclusion criteria, and it is not possible to directly compare their results to track temporal changes.

Here, we develop and apply a semi-automated natural language processing (NLP) framework to categorize a funding landscape at user-specified levels of granularity. We apply this method at the levels of 15 and 60 research topic clusters to nearly 6,000 NCI research grants awarded to radiation oncology or radiology over 21 years. We track funding changes over time by topic. Manual validation on a small subset of the data confirmed the quality of the cluster assignments to be similar to human assignment.

Our code is publicly available and runs efficiently on standard laptop computers. To our knowledge, this is the first study that has interrogated multi-decade funding by an NIH institute using machine learning and natural language processing (NLP). This method provides a comprehensive overview of 1.9 billion dollars of NCI funding over 21 years and is extensible to other fields of study.

## Methods and Materials

### Dataset

From NIH RePORTER, we retrieved the 19,945 NIH grants funded from FY 2000 through 2020 awarded to principal investigators (PI) with primary affiliation in departments of radiation oncology or diagnostic radiology (department field of "Radiation – Diagnostic/Oncology"). 1,345 grants with empty abstracts or abstracts shorter than 50 word-tokens after preprocessing were removed. This was narrowed to the 7,508 grants from the NCI, and only R-type grants (excluding R25 education program grants) were considered, leaving a final set of 5,874 grants (**Figure 1**; **Supplementary Table S1**). Affiliated department categorization is only available for medical school departments, so our dataset only includes grants awarded to departments with affiliated medical schools[2,3].

[1] https://www.cancer.gov/about-nci/budget/fact-book/archive/2020-fact-book.pdf

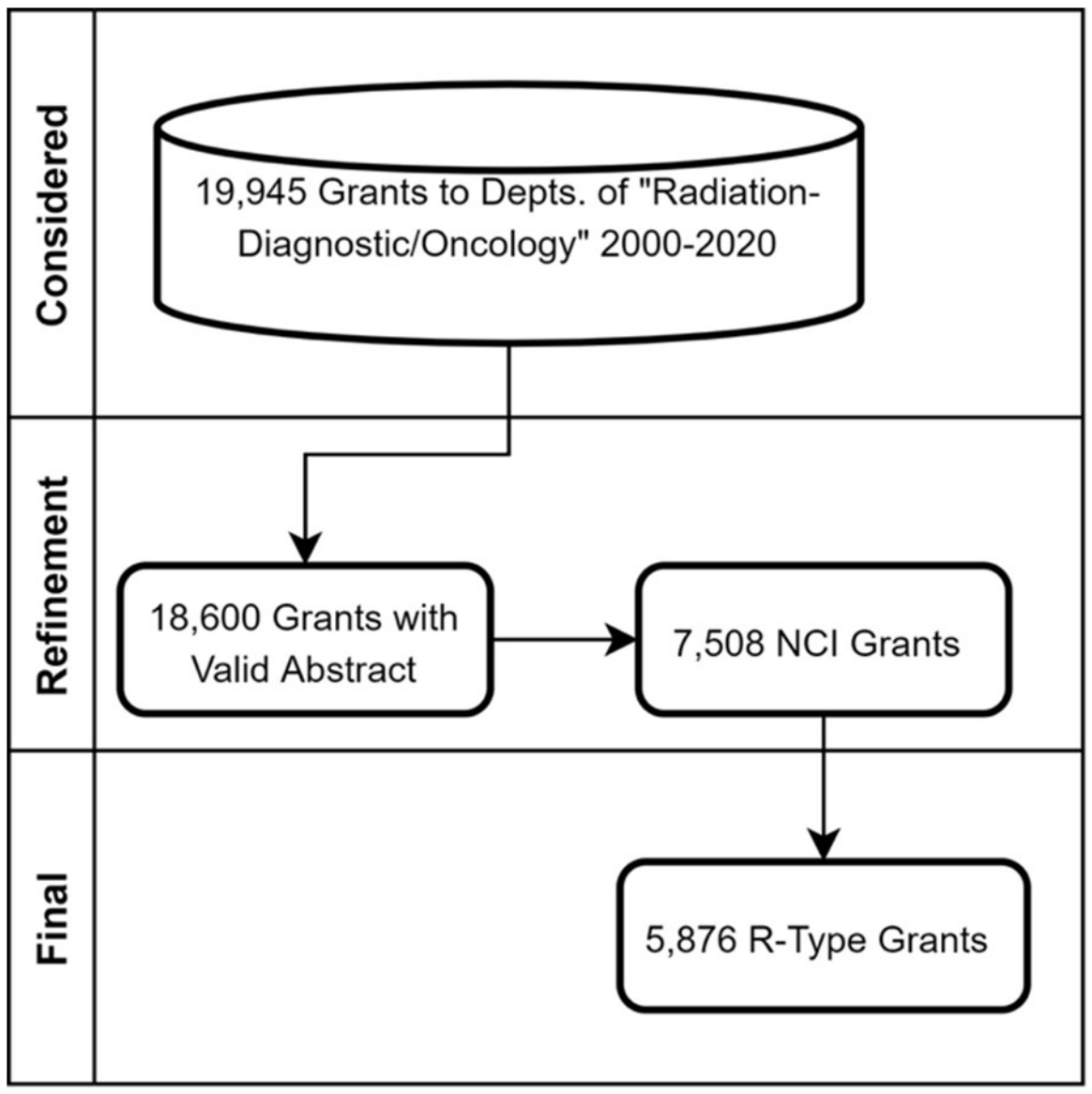

**Figure 1**: A database of 19,945 grants awarded to the Department of "Radiation-Diagnostic/Oncology" from FY 2000 to 2020 was narrowed down to 18,600 grants with valid abstracts and refined to 7,508 NCI grants. We further streamlined the dataset to include only 5,876 R-Type Grants (excluding R-25 grants). This final grant corpus includes 1790 unique abstracts with 4086 renewals.

Data preparation

Preprocessing in Python version 3.8.10 consisted of word tokenization (via NLTK library), stop-word removal (via Sci-Kit Learn library) and removal of tokens that appeared less than 50 times in the dataset.

After preprocessing, abstracts were converted to weighted 200-dimensional feature vectors using a 2-step process. (1) Pretrained word token vectors from biomedical and clinical text were used to represent each token[8]. These 200-dimensional vectors have been trained on PubMed Central (28,714,373 documents) and a database of de-identified EHR records (2,083,180 clinical notes) from MIMIC III[9]. (2) The product of term frequency and inverse document frequency (TF-IDF) was then used as a weighting function to average multiple word vectors to a single 200-dimensional vector representing each abstract[10] (**Supplementary Text S2**).

Hierarchically stabilized k-means clustering

Our method is based on a variant of k-means clustering, which is a classic unsupervised learning model used to group similar, unlabeled data together into "k" clusters where k is chosen by the user. We used Ward agglomerative clustering, a hierarchical method, to generate initial centroids for k-means clustering, which was used to generate final cluster centroids. This combined approach gives deterministic centroids and has better performance than either method alone (**Supplementary Figure S2**). We assessed clustering performance using silhouette score[11] at different values of k (**Supplementary Figure S3**). Clustering performance was visualized via t-SNE[12] (**Supplementary Figure S4**). Our source code is available at https://github.com/kang-lab/Rad_clustering.

Manual validation of clustering

Reviewers manually evaluated clustering performance. Two medical students (PB, MN), one radiation oncology resident (JT) and one radiation oncologist (JK) reviewed 300 abstracts, which represented 5% of the 5,876 grants. Each of the four reviewers rated 100 abstracts with 25% overlap to assess interrater agreement, for a total of 400 manual reviews with 100 reviewed by 2 reviewers. In a review, a grant was represented by its abstract and title, and the reviewer was asked to pick the most appropriate cluster from five options—one of which was the cluster assigned this grant by our model, and the other four were randomly chosen among the remaining clusters. Each cluster was represented by its top 10 words by overall TF-IDF score. An example is shown in **Supplementary Figure S5**.

Research topic naming

After manual validation, we assigned names to clusters to create interpretable research topics. Research topics were named based on the important tokens by TF-IDF values in their respective cluster, as well as the 10 unique closest abstracts by Euclidean distance to each cluster centroid. Research topic naming underwent five rounds of iterative review and editing by all authors, which included medical students (MN, PB), radiation oncology residents (JT, AA), a radiation oncologist (JK), a diagnostic radiologist (DC), and a diagnostic radiology medical physicist (PK) to achieve final consensus.

Funding trends

After clustering all grant abstracts into their respective research topic, we then use corresponding NIH RePORTER data to calculate the total funding for each topic and absolute change over a 21-year period from FY 2000-2020. Research topics were then labeled from 1 to 15 (k=15) or 60 (k=60), with topic #1 corresponding to the largest increase in funding and topic #15 or #60 corresponding to the largest decrease in funding.

## Results

Our final dataset comprises abstracts from 5,876 R-type grants from NCI to departments of "Radiation-Diagnostic/Oncology" from FY 2000-2020 (**Figure 1**; **Supplementary Table S1**). These abstracts had a mean length of 231 word-tokens per abstract and the corpus had a total vocabulary of more than 25,000 unique word tokens after preprocessing.

### Clustering

The number of clusters or cluster centroids (hyperparameter k) represents the desired number of topics or research areas for the clustering algorithm to assign grants. Multi-metric elbow plot heuristics suggest approximately 13 clusters could optimally represent this dataset before clustering performance degrades (**Supplementary Figure S3**), and thus we use 15 clusters to represent one view of the funding landscape. Our preliminary data suggested that 15 clusters did not provide the desired level of granularity desired to capture the breadth of research funding. Thus, we ultimately chose two levels of clustering granularity—15 and 60 clusters—to capture a less and a more granular view of the funding landscape, respectively. We will refer to 15 and 60 cluster results as k=15 and k=60, respectively.

### Visualization

t-distributed Stochastic Neighbor Embedding (t-SNE), a non-linear dimensionality reduction technique[12], was used to visualize the clustered 200-dimensional abstract encodings in a 2-D space. Surprisingly, plotting the k=15 and k=60 centroids in two dimensions revealed that the grant embeddings in the 2-D space can be represented by 2 axes: (1) physics/biology and (2) diagnostics/therapeutics. With k=15, the physics/biology extremes include *Radiotherapy & CBCT Technology* and *Novel Imaging Systems* on the physics side and *DNA damage & repair* and *Cancer genetics* on the biology side (**Figure 2a**). On the diagnostics/therapeutics axis, the diagnostics extreme includes *Automated detection* and the therapeutics extreme includes *Molecular therapies.* These patterns hold in the k=60 plot (**Figure 2b**), though with the notable exceptions of Clusters 20 and 47 representing *Organ-specific ablation and functional imaging* and *Ablation, oxygenation, novel MRI methods*, respectively, being placed on the diagnostics side; this notable discrepancy will be addressed in the Discussion.

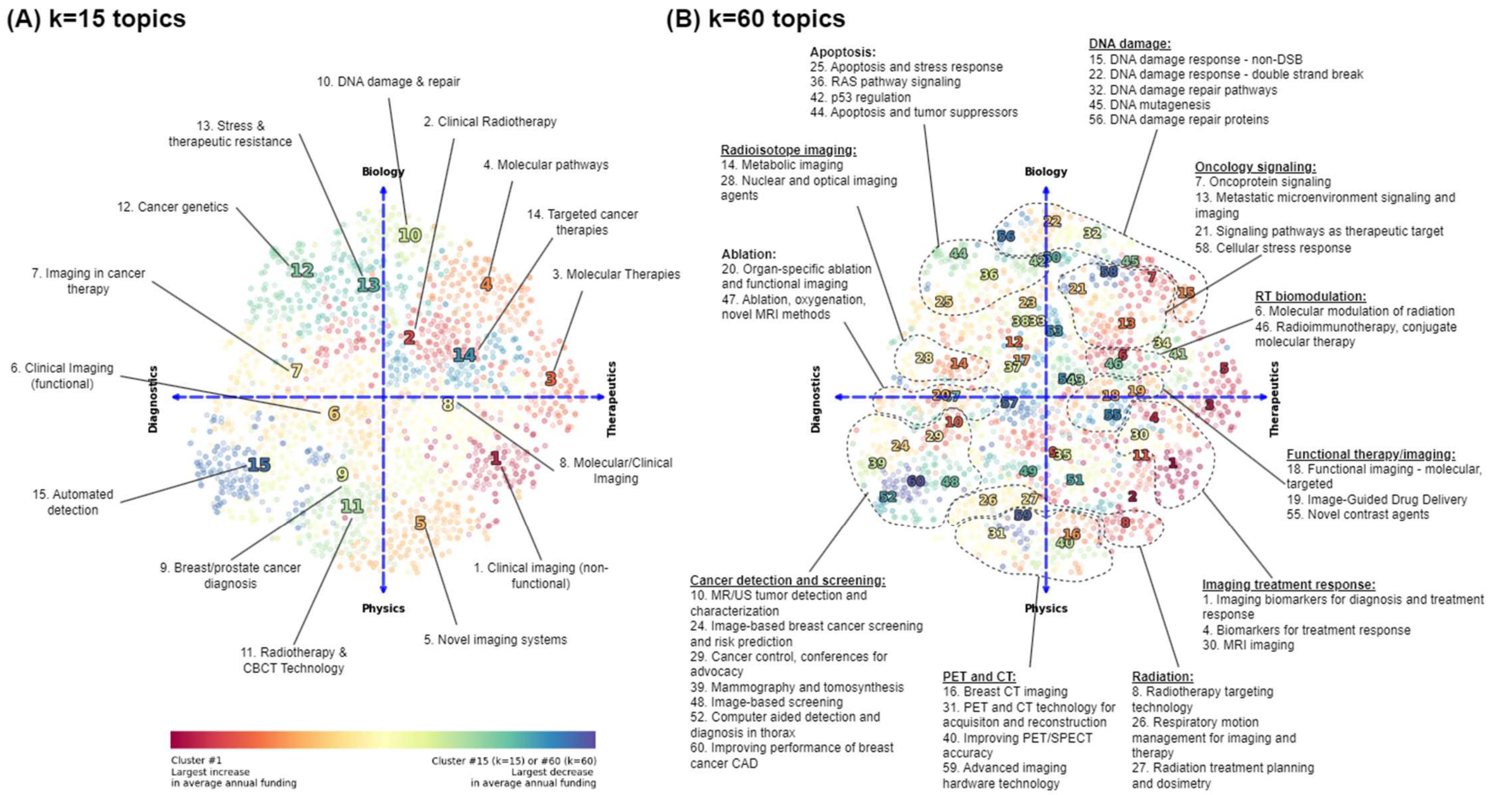

**Figure 2**: Two-dimensional visualization in t-SNE space of 5,874 NCI R-type grants awarded to departments of diagnostic radiology or radiation oncology in FY 2000-2020. Grant embeddings were assigned to (a) k=15 or (b) k=60 clusters. Each circle represents 1 grant. The clusters are both ranked and color-coded according to their absolute funding growth rate with cluster 1 and red corresponding to greatest increase in funding from FY 2000-2020. For k=60, clusters sharing topic overlap are manually separated with dotted lines. See **Figure 4** for full legend of cluster names for k=60 topics.

## Manual validation

For the 400 manual grant reviews, the Cohen's kappa agreements between reviewers and the clustering model were 0.53±0.07 and 0.71±0.06 (mean ± standard deviation) reflecting moderate and substantial agreement for k=15 and k=60, respectively[13] (**Supplementary Figure S6**). For k=15, grants that are further from their respective centroid are more likely to have human disagreement about which topic they belong to (**Supplementary Figure S7**).

For the 100 manual grant reviews that were read by 2 reviewers, the Cohen's kappa interrater agreement between 2 reviewers was 0.68±0.09 for k=15 and 0.71±0.06 for k=60 (mean ± standard deviation), both of which are considered substantial agreement[13]. Since Cohen's kappa scores between human evaluators are similar to human agreement with k=60 clustering, we conclude that our model performs similarly at assigning documents to clustering topics as a human at higher granularity.

## Funding trends

By clustering this corpus into k=15 and k=60 research topics, many funding trends emerged. Funding trends at k=15 and k=60 granularity are incorporated into the clustering t-SNE visualization (**Figure 2**). The abstracts on the therapeutic semicircle have a faster overall growth rate than the diagnostic side (**Supplementary Table S2**).

Consensus topic names are shown in **Figure 3** for k=15. The top 3 k=15 research topics with highest increase in funding are *Clinical imaging (non-functional)*, *Clinical Radiotherapy* and *Molecular Therapies*, which all had an absolute growth of >$530,000 per year, while *Stress & therapeutic resistance*, *Targeted cancer therapies*, *Automated detection* had the highest annual decreases of -$74,293, -$158,854, -$178,090, respectively (**Figure 3a**). Notably, *Automated detection* received the most funding in 2000 (roughly $9 million) but its total amount of

funding decreased to approximately $7.5 million in 2020 while *Clinical imaging (non-functional)* grew from $0.4 million in 2000 to $16.4 million in 2020 (**Figure 3b**).

Consensus topic names are shown in **Figure 4** for k=60 topics. The top 3 k=60 research topics with the highest increase in funding are *Imaging biomarkers for diagnosis and treatment response*, *Informatics software for treatment decision support* and *Radiopharmaceuticals* which all grew by >$218,000 per year. *Cellular stress response*, *Advanced imaging hardware technology,* and *Improving performance of breast cancer CAD* saw the largest decrease in funding (-$117,078, -$121,618, -$175,622 per year, respectively) (**Figure 4a**).

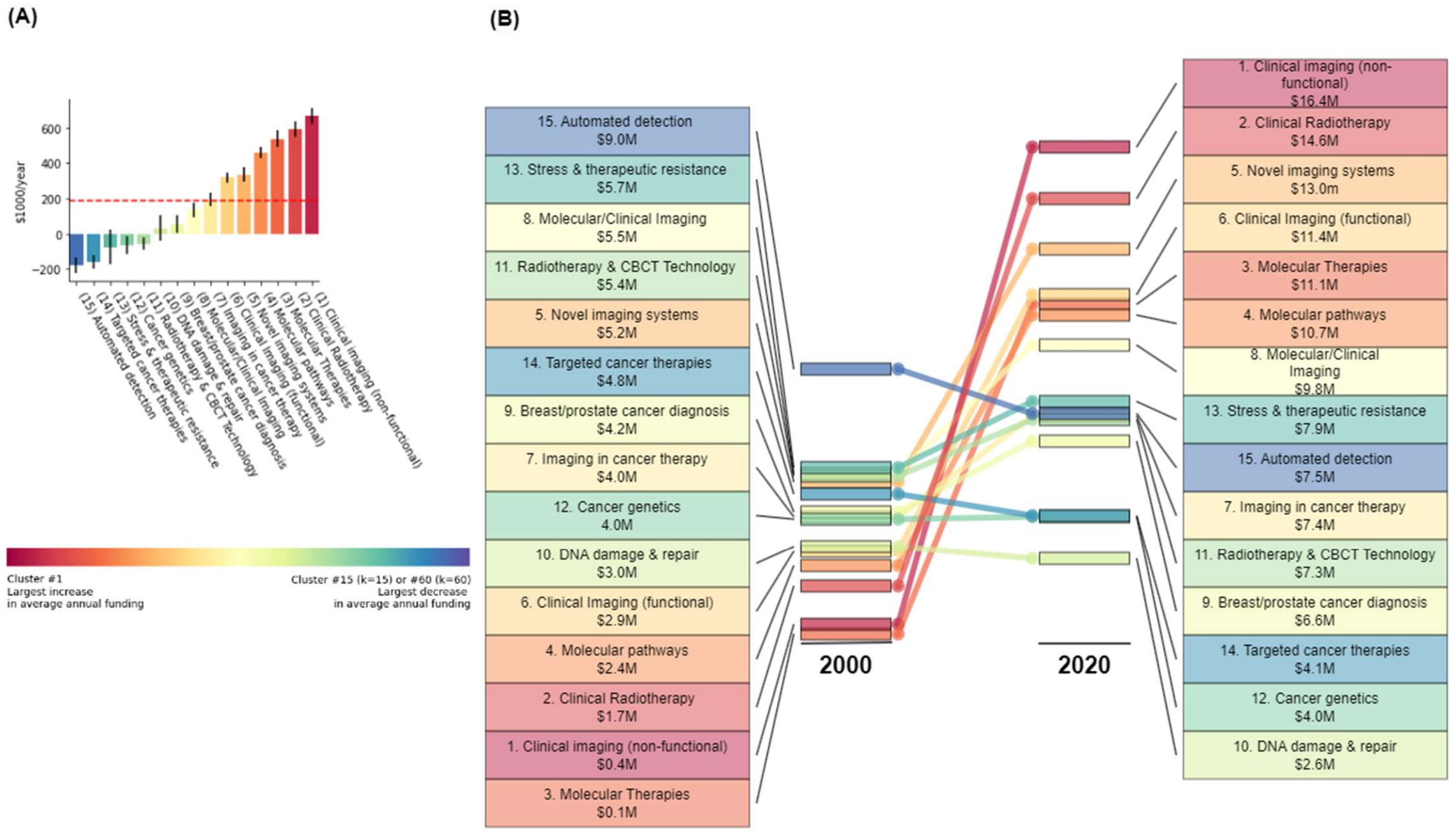

**Figure 3**: Average annual growth and total funding analysis for granularity-level k=15 estimated research topics at FY 2000 and FY 2020. Topics are numbered and color-coded according to their absolute growth rate, with topic 1 and red corresponding to the greatest increase and topic 15 and blue corresponding to the greatest decrease in funding. (a) Growth rate per year with the dotted red line depicting average growth rate of all research topics. (b) Total funding amount, to the nearest 0.1 million, of all k=15 research topics in 2000 and 2020.

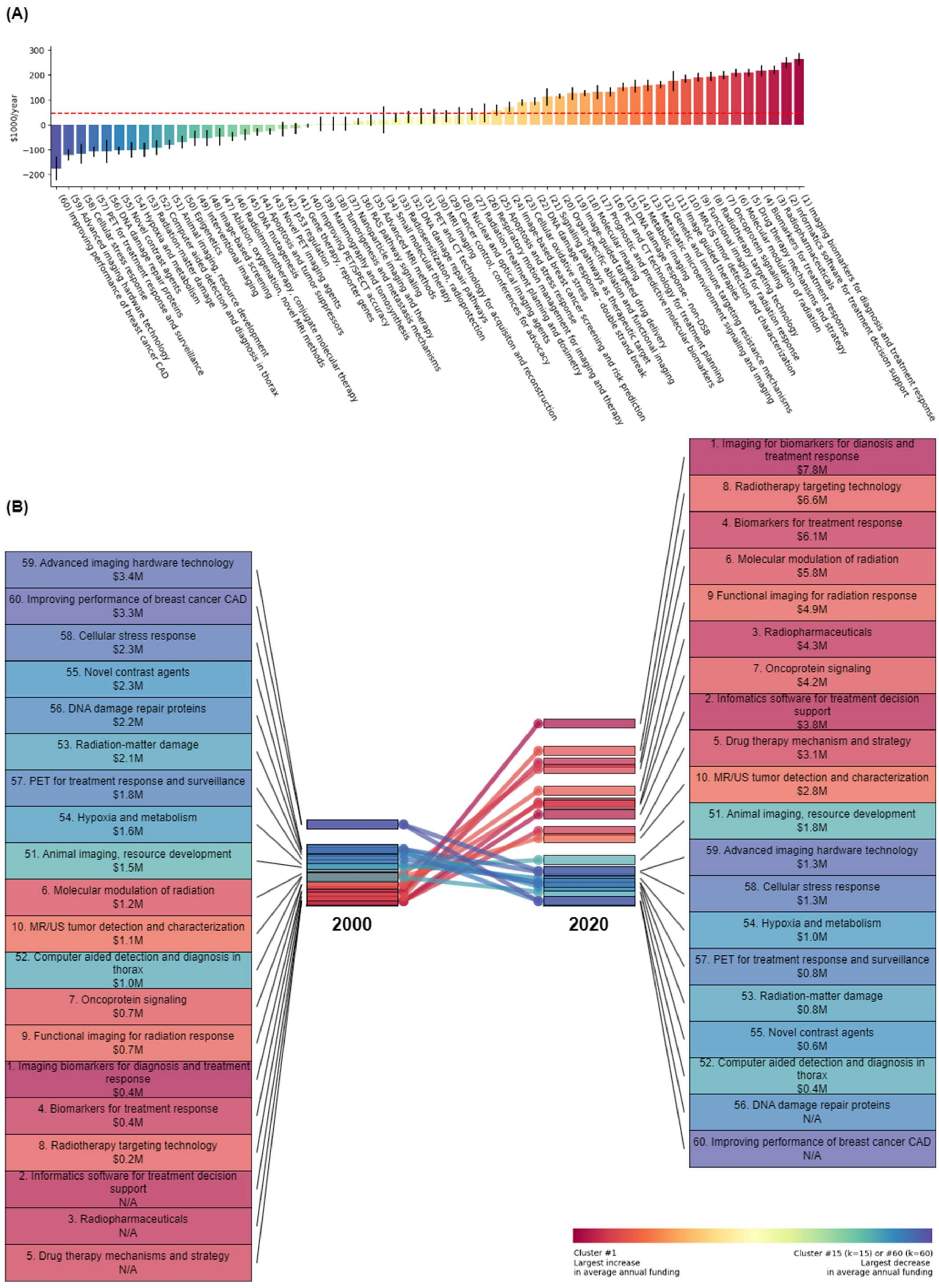
(A)
300
200
100
0
−100
−200
$1000/year
(60) Improving performance of breast cancer CAD
(59) Advanced imaging hardware technology
(58) Cellular stress response
(57) PET for treatment response and surveillance
(56) DNA damage repair proteins
(55) Novel contrast agents
(54) Hypoxia and metabolism
(53) Radiation-matter damage
(52) Computer aided detection and diagnosis in thorax
(51) Animal imaging, resource development
(50) Epigenetics
(49) Interventional imaging
(48) Image-based screening
(47) Ablation, oxygenation, novel MRI methods
(46) Radioimmunotherapy, conjugate molecular therapy
(45) DNA mutagenesis
(44) Apoptosis and tumor suppressors
(43) Novel PET imaging agents
(42) p53 regulation
(41) Gene therapy, reporter genes
(40) Improving PET/SPECT accuracy
(39) Mammography and tomosynthesis
(38) Tumorigenesis and metastasis mechanisms
(37) Nanoparticle imaging and therapy
(36) RAS pathway signaling
(35) Advanced MRI methods
(34) Small molecular therapy
(33) Radiosensitization, radioprotection
(32) DNA damage repair pathways
(31) PET and CT technology for acquisiton and reconstruction
(30) MRI imaging
(29) Cancer control, conferences for advocacy
(28) Nuclear and optical imaging agents
(27) Radiation treatment planning and dosimetry
(26) Respiratory motion management for imaging and therapy
(25) Apoptosis and stress response
(24) Image-based breast cancer screening and risk prediction
(23) Cellular oxidative stress
(22) DNA damage response - double strand break
(21) Signaling pathways as therapeutic target
(20) Organ-specific ablation and functional imaging
(19) Image-guided targeted drug delivery
(18) Molecular imaging
(17) Prognostic and predictive molecular biomarkers
(16) PET and CT technology for treatment planning
(15) DNA damage response - non-DSB
(14) Metabolic imaging
(13) Metastatic microenvironment signaling and imaging
(12) Genetic and immune targeting resistance mechanisms
(11) Image guided therapies
(10) MR/US tumor detection and characterization
(9) Functional imaging for radiation response
(8) Radiotherapy targeting technology
(7) Oncoprotein signaling
(6) Molecular modulation of radiation
(5) Drug therapy mechanisms and strategy
(4) Biomarkers for treatment response
(3) Radiopharmaceuticals
(2) Informatics software for treatment decision support
(1) Imaging biomarkers for diagnosis and treatment response
(B)
59. Advanced imaging hardware technology $3.4M
60. Improving performance of breast cancer CAD $3.3M
58. Cellular stress response $2.3M
55. Novel contrast agents $2.3M
56. DNA damage repair proteins $2.2M
53. Radiation-matter damage $2.1M
57. PET for treatment response and surveillance $1.8M
54. Hypoxia and metabolism $1.6M
51. Animal imaging, resource development $1.5M
6. Molecular modulation of radiation $1.2M
10. MR/US tumor detection and characterization $1.1M
52. Computer aided detection and diagnosis in thorax $1.0M
7. Oncoprotein signaling $0.7M
9. Functional imaging for radiation response $0.7M
1. Imaging biomarkers for diagnosis and treatment response $0.4M
4. Biomarkers for treatment response $0.4M
8. Radiotherapy targeting technology $0.2M
2. Informatics software for treatment decision support N/A
3. Radiopharmaceuticals N/A
5. Drug therapy mechanisms and strategy N/A
2000
2020
1. Imaging for biomarkers for dianosis and treatment response $7.8M
8. Radiotherapy targeting technology $6.6M
4. Biomarkers for treatment response $6.1M
6. Molecular modulation of radiation $5.8M
9 Functional imaging for radiation response $4.9M
3. Radiopharmaceuticals $4.3M
7. Oncoprotein signaling $4.2M
2. Infomatics software for treatment decision support $3.8M
5. Drug therapy mechanism and strategy $3.1M
10. MR/US tumor detection and characterization $2.8M
51. Animal imaging, resource development $1.8M
59. Advanced imaging hardware technology $1.3M
58. Cellular stress response $1.3M
54. Hypoxia and metabolism $1.0M
57. PET for treatment response and surveillance $0.8M
53. Radiation-matter damage $0.8M
55. Novel contrast agents $0.6M
52. Computer aided detection and diagnosis in thorax $0.4M
56. DNA damage repair proteins N/A
60. Improving performance of breast cancer CAD N/A
Cluster #1
Largest increase in average annual funding
Cluster #15 (k=15) or #60 (k=60)
Largest decrease in average annual funding

**Figure 4**: Average annual growth and total funding analysis for granularity-level k=60 estimated research topics. Topics are numbered and color-coded according to their absolute growth rate, with topic 1 and red corresponding to the greatest increase and topic 60 and blue corresponding to the greatest decrease in funding. (a) Growth rate per year of all 60 topics with the dotted red line depicting the average growth rate of all k=60 research topics. (b) Total funding amount, to the nearest 0.1 million, of the top 10 research topics that experienced the greatest increase or decrease in funding in 2000 or 2020. Research topics that did not contain awarded grants in 2000 or 2020 were labeled with "N/A" after their respective topic name.

Emergence of new funding k=60 topics

Further scrutiny reveals complete disappearance and emergence of k=60 topics over FY 2000-2020 (

**Figure 5**), which can be partially appreciated on **Figure 4b.** k=60 clusters that disappeared completely include *DNA damage repair proteins*, *Improving performance of breast cancer CAD, Epigenetics*, and *DNA mutagenesis* after FY 2010, 2015, 2016, 2018, respectively. Conversely, many new research topics have emerged such as *Drug therapy mechanisms and strategy*, *Image guided therapy*, *Genetic and immune targeting resistance mechanisms, Image-guided targeted drug delivery, Informatics software for treatment decision support*, *Metabolic Imaging,* and *Radiopharmaceuticals*. Notably, since its first appearance in 2004, *Informatics software for treatment decision support* has been the fastest-growing cluster.

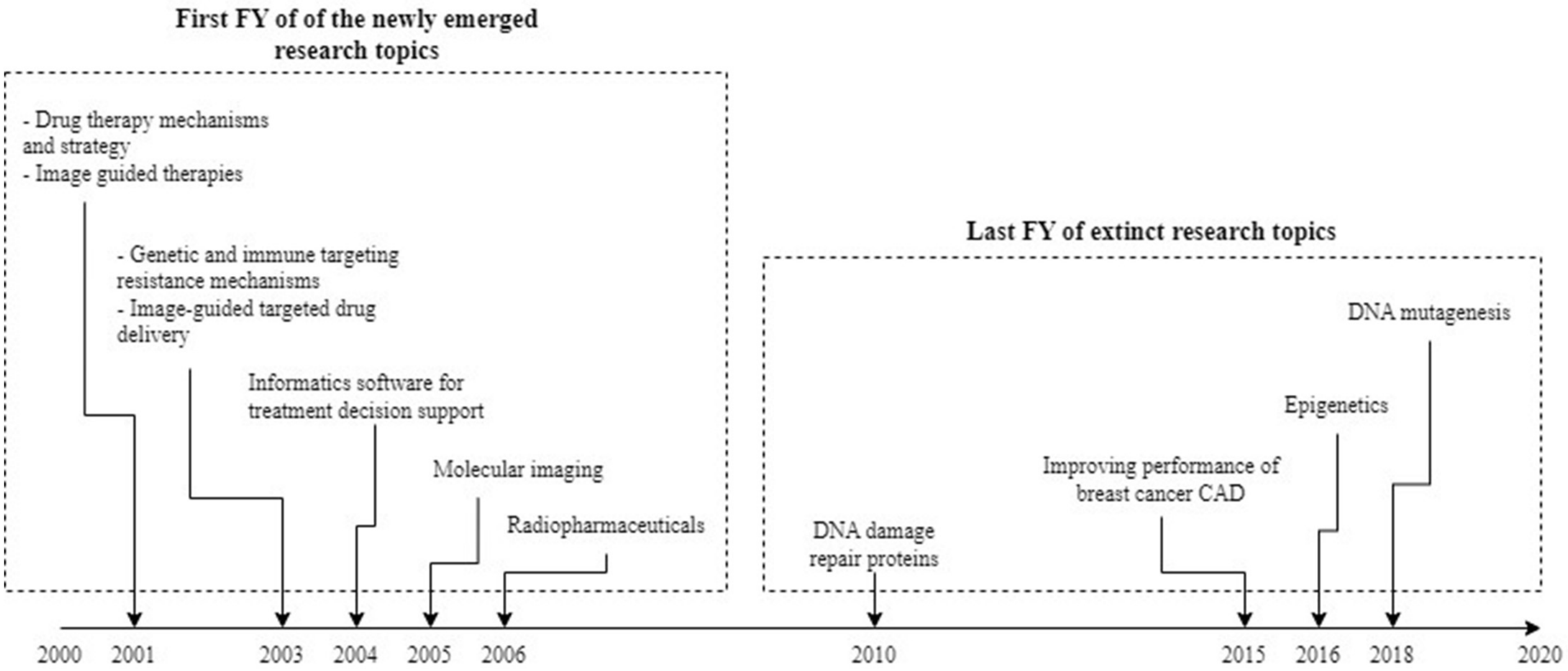

**Figure 5**: Newly emerged and extinct research topics funded by NCI to departments of diagnostic radiology or radiation oncology from 2000-2020. These results do not indicate that there was no research in these topics before or after these dates, but that collective research did not rise to the level to be detected at a k=60 granularity.

## Discussion

In this study, we combined NLP, unsupervised machine learning, and subject matter expertise to identify clusters of research topics, which we use to estimate trends in NCI funding for the radiological sciences over the past two decades. We demonstrated the validity of our results in several ways: human-human interrater agreement, human-computer interrater agreement, and visualization of dimension-reduced data. Manual validation showed that our model assigned documents to clusters similarly as well as humans when presented with multiple choices. Interpretation of cluster visualization shows that representing grants as word embeddings pre-trained on biomedical text retain meaningful semantic information about research topics, and visualization of these embeddings reveals high-level research axes. We observed that increased subject matter expertise improved the quality of name assignment to the k=60 clusters as they are more granular and require more in-depth knowledge than k=15. We also noticed that k=60 clusters further from origin have objectively higher clustering quality metrics and were more easily defined than more central clusters (**Figure 2b**).

Our current method relies on the user to pre-specify the number of desired topics to select the desired granularity; the true number of topics is an unknown and likely subjective quantity. There is evidence of overlapping topics at k=60: the *DNA damage response - non-DSB* and *DNA damage response – double strand break* topics are similar and occupy the same space on the t-SNE plot (**Figure 2b**). The k-means clustering algorithm generally tries to find well-balanced cluster sizes (**Supplementary Figure S1**) which suggests that clusters could be grouped into similar meta-clusters to better capture more general research domains, such as meta-cluster *DNA damage* containing 5 related clusters in **Figure 2b**, including the two aforementioned k=60 clusters.

Upon cross-referencing with the k=15 t-SNE visualization (**Figure 2a**), we see the region containing the k=60 *DNA damage* meta-cluster overlaps with the k=15 *DNA damage & repair* and a part of the *Molecular pathways* topic. Similar corresponding areas of research between the k=15 clusters and the k=60 meta-clusters can be seen across a breadth of research areas, providing another form of validation that our framework captures biomedical semantic representation of NCI grants research.

Spatial incongruencies

While the observed t-SNE axes fit well with most centroid names, our method placed the k=60 clusters for *Organ-specific ablation and functional imaging* and *Ablation, oxygenation, novel MRI methods* near the diagnostics end of the diagnostics-therapeutics axis despite ablation being a therapy (**Figure 2b**). Since word tokens are weighted by their relative uniqueness to the grant via TF-IDF (**Supplementary Text S2**) and our method assumes that the semantic meaning (i.e., encoding) of a grant is compromised of the collective meaning of its individual words, we suspect the high proportion of imaging-related word-tokens (e.g. "MRI" or "ultrasound") are suppressing the therapy-related tokens (e.g. "PDT" and "ablation") in grants within these clusters. Despite this spatial incongruency on t-SNE, our method appears to have correctly clustered the grants related to ablation together into the *Ablation* meta-cluster for k=60 (**Figure 2b**).

Interpretation challenges

While topic names underwent iterative revisions by subject matter experts, they are ultimately subjective and are intended to estimate a curated representation of a research domain in high dimensional semantic embedding space. Despite this limitation, manual validation suggests that clustering performance was similar to human interrater agreement in categorizing grant abstracts for k=60 (**Supplementary Figure S6**).

Our methodology was designed to run without high performance computing and thus does not consider contextual definitions or word order. The expected benefit of contextual and positional embeddings is low given matched domains of the pre-trained biomedical embeddings and grants corpus though we plan to investigate this benefit in future work.

Interpretation of growth trends by topic (**Figure 3, Figure 4**) must consider that there is no causal link implied between increased in topic funding and increasing interest by NCI. Unfunded grants abstracts are not publicly available, and thus we cannot determine whether funding trends are a cause or an effect of evolving interest by NCI. Other unmeasured confounders may the evolution of terminology and the consolidation of research fields as new techniques mature. One example from k=60 is the rapid growth of *Informatics software for treatment decision support* concurrent with concomitant shrinkage of *Improving performance of breast cancer CAD* and *Computer aided detection and diagnosis in the thorax*. Computer aided diagnosis (CAD) is a type of decision support/AI, but research in this area predated the current surge of interest in AI.

Our model assigns each grant abstract to the closest centroid by design, a framework that has been used for similar problems[14–16], which can lead to seemingly incorrect assignments particularly for clusters further from the center (**Supplementary Figure S7**). A soft clustering method that allows for multiple topic assignments could be considered[17–20] to better represent the complexity of research though would also increase model complexity. Whether soft clustering would increase or decrease interpretability is not clear as adding additional complexity may make visualization more challenging.

Comparison with other work

Our clustering approach for grant topic modeling differs from traditional Latent Dirichlet Allocation (LDA) methods by using word embeddings to pre-encode biomedical knowledge[8,10] and is simpler than the embedded topic model method[21], which was also developed to generate document embeddings from word embeddings. At the time when we were developing our methods, similar methods were being proposed[21,22]. For future work, we plan to finetune our data preprocessing, use document vectors[23–25], and consider stacked encoding large language models that may help overcome multi-context issues (e.g., "ultrasound" being used for both diagnosis and therapy).

## Conclusion

We extracted research topics that the NCI funded for the radiological sciences from 2000 to 2020 at two levels of granularity using a text corpus of R-type grant abstracts. Our NLP framework combines unsupervised machine learning with subject matter expert review to extract research topic names. We reported the estimated funding growth and shrinkage of these topics over a 21-year period. Visualization of our document semantic representations unexpectedly revealed orthogonal physics-biology and diagnostics-therapeutics axes allowing multi-scale interpretation of the results. Our framework scales well, is computationally inexpensive, is comparable to human topical assignment in quality, and has a similar basis to other approaches in the NLP community. We believe our framework represents a novel, viable approach to longitudinal topic analysis of grant funding abstracts and could be applied to other research areas.

**Supplement**

Prior funding research

**Supplementary Text S1**: history of funding research in radiation oncology by ASTRO

In 2013, Steinberg et al. described 1 year of funding distribution across grant mechanisms, career stage, and 3 categories: physics, biology, and clinical investigation. They used manual internet searches of primary investigators to differentiate between radiology and radiation oncology grants, concluding that radiation oncology may be underfunded by the NIH. No specialty-labeled list of faculty or NIH projects is publicly available for further analysis.

To better understand the funding landscape, a 2011 ASTRO task force used large scale surveys and interviews to understand research interests and identify radiation oncology-specific NIH awards. They then created 17 biology categories and outlined 10 separate topic categories as future directions (Wallner 2014). The creation of these topics was subjective and covered three years of NIH awards, meant to act as a snapshot.

In 2017, Yu et al. used surveys to identify 1,847 funded and unfunded applications that were specifically related to radiation oncology. The authors used titles to assign grants to 5 topic categories outlined in the 2017 ASTRO research agenda. These categories are different from those in the Wallner study, making direct comparison impossible.

**Supplementary Text S2**: topic modeling background

Our goal was also to generate document embeddings from word embeddings to better capture the nuances of a word. These word vectors have been shown to represent semantic meaning of words, which in this work allows us to better represent the meaning of grant abstract text. For example, the word vectors of "cancer" and "tumor" are closer together in high dimensional space than the word vector of "cancer" and "paper". To encode grant abstracts, we used word vectors in conjunction with the TF-IDF metric of word importance. TF-IDF is a common metric of word significance to the meaning of a document and is the product of the number of times a word appears in a document and the inverse of its frequency in the corpus. Incorporating TF-IDF as a weighting function reduces the contribution of vectors representing words like "the" or "a," and this has been shown to improve clustering performance. The combination of pretrained word vectors with their TF-IDF value allows our model to effectively represent the research topics of an abstract.

Data statistics

**Supplementary Table S1**: Breakdown of funding mechanisms in our corpus. R01 grants comprise (79%) of the corpus.

| R-type grants | Number of abstracts |
|---|---|
| R00 | 39 |
| R01 | 4647 |
| R03 | 45 |
| R13 | 37 |
| R21 | 802 |
| R24 | 50 |
| R29 | 46 |
| R33 | 93 |
| R35 | 36 |
| R37 | 49 |
| R38 | 1 |
| R50 | 16 |
| R56 | 5 |
| RC1 | 8 |

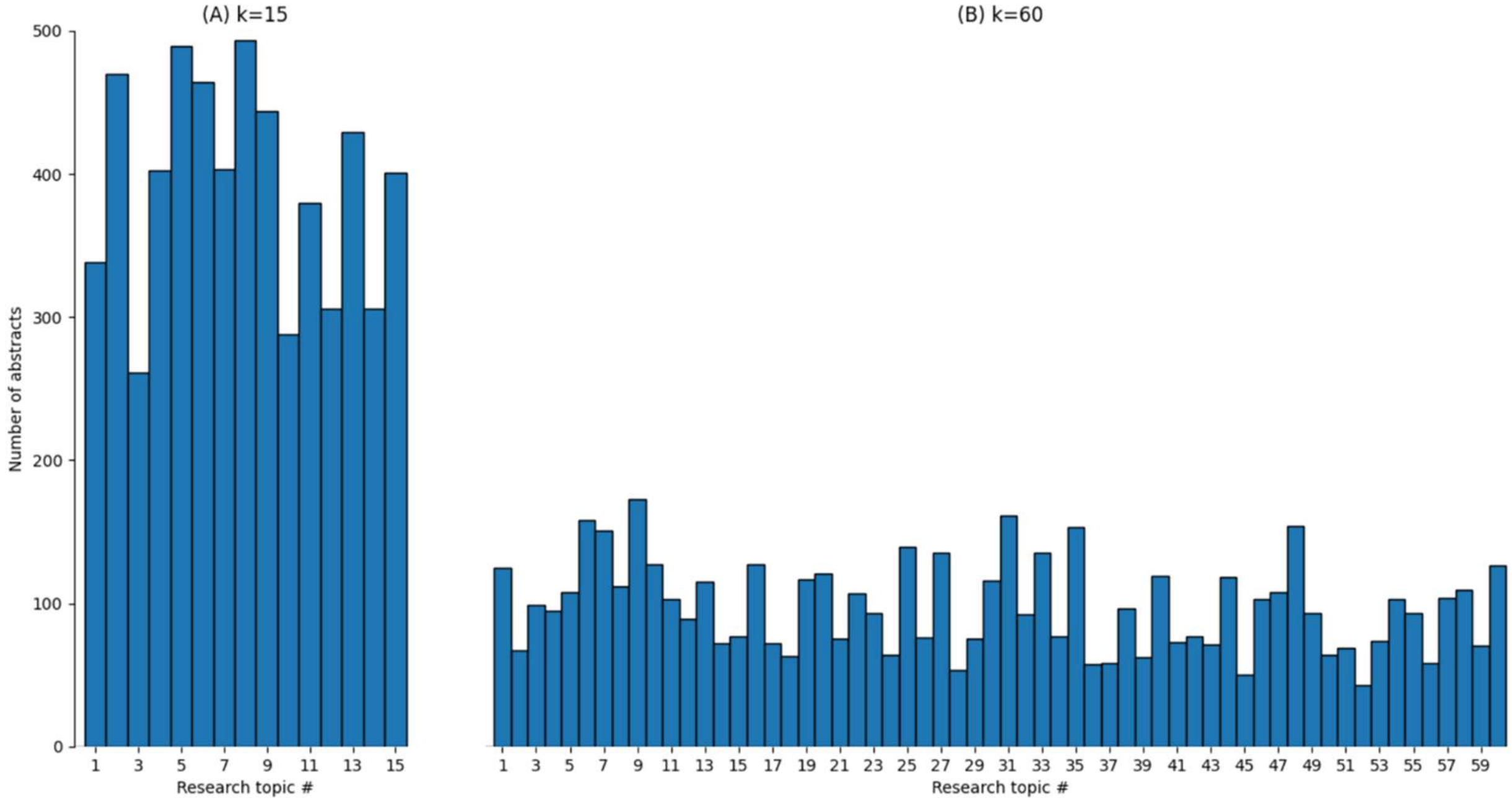

**Supplementary Figure S1**: Numbers of grants per topic at k=15 and k=60.

Hierarchically stabilized versus original k-means

To evaluate our clustering method vs established methods, we measured the silhouette score of clustering over a range of k-values from 2 to 80. k-means and Agglomerative (Ward) clustering were implemented using Sci-kit Learn with default hyperparameters except the number of k-means initializations was increased to 100. Although the silhouette scores of k-means are only slightly lower than those of the combination method, k-means does not produce consistent results. As a result, we decided to use hierarchical clustering to generate stable centroids to initialize k-means. This extra step reduces the randomness of the traditional k-means.

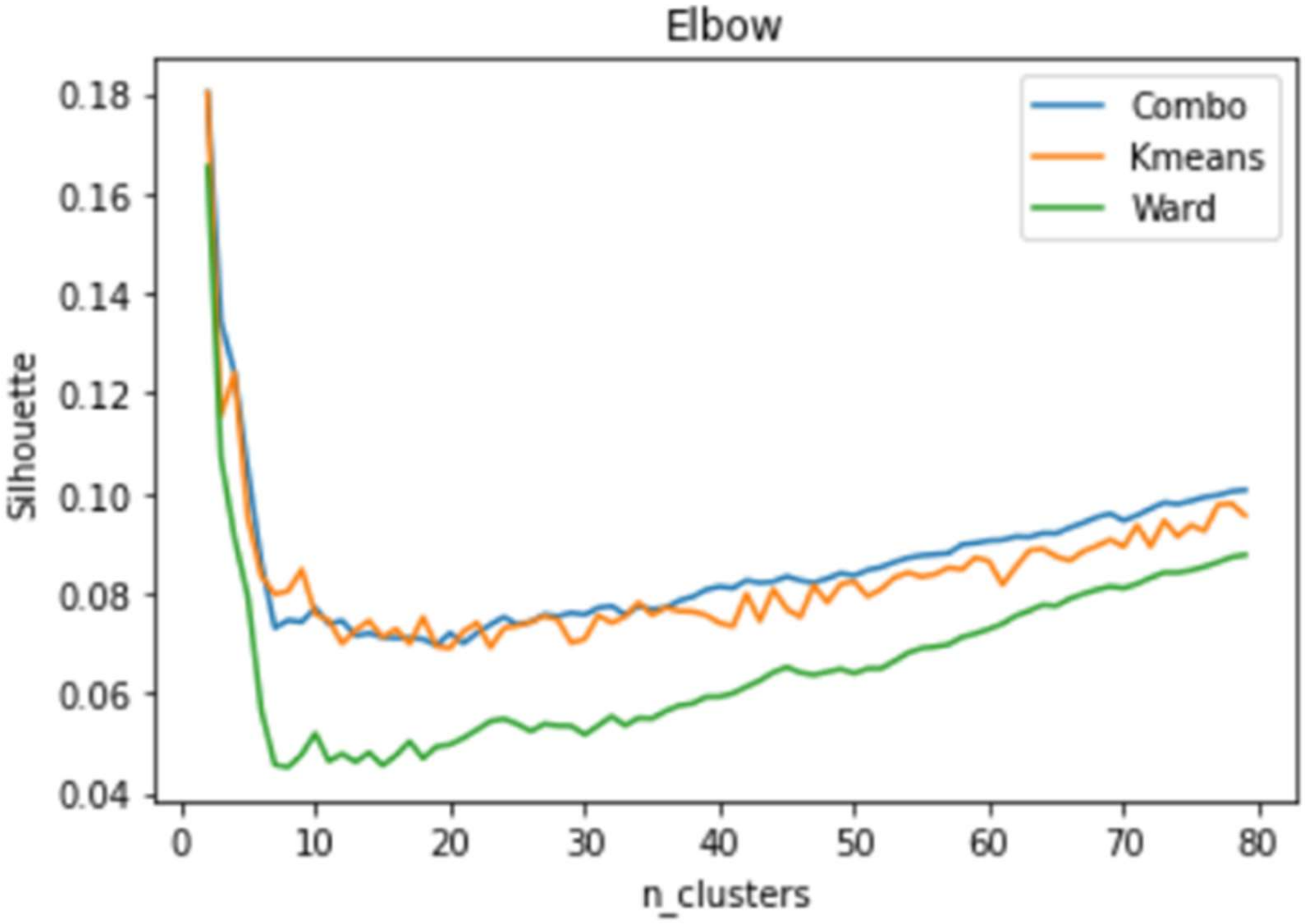

**Supplementary Figure S2**: Clustering quality as defined by silhouette score of k-means, agglomerative, and hierarchically stabilized k-means at different numbers of clusters.

Number of clusters

The number of clusters (k) remains the most subjective hyperparameter because of a lack of a ground truth value. In this study, we employed 2 methods to define the number of clusters (k) – the elbow method and expert recommendation. The elbow method defines the k value by plotting the rate of change of inertia as a function of k. Inertia computes the sum of square distance of each datapoint in a cluster to the cluster centroid so the smaller the inertia value, the better are the clustering quality. As demonstrated using our data, inertia is largest at k=1 and as we increase the number of clusters to 2, inertia drops sharply because unrelated datapoints are no longer forced to be in the same cluster. This sharp drop in inertia continues until k exceeds the true number of clusters and at this point, adding more clusters does not provide significant benefit to data clustering. As a result, any k > 12 clusters can sufficiently describe the number of topics in this dataset. After collecting expert recommendations, we decided to analyze research trends at k=15 and k=60 to capture the funding trends at both low and high level of granularity.

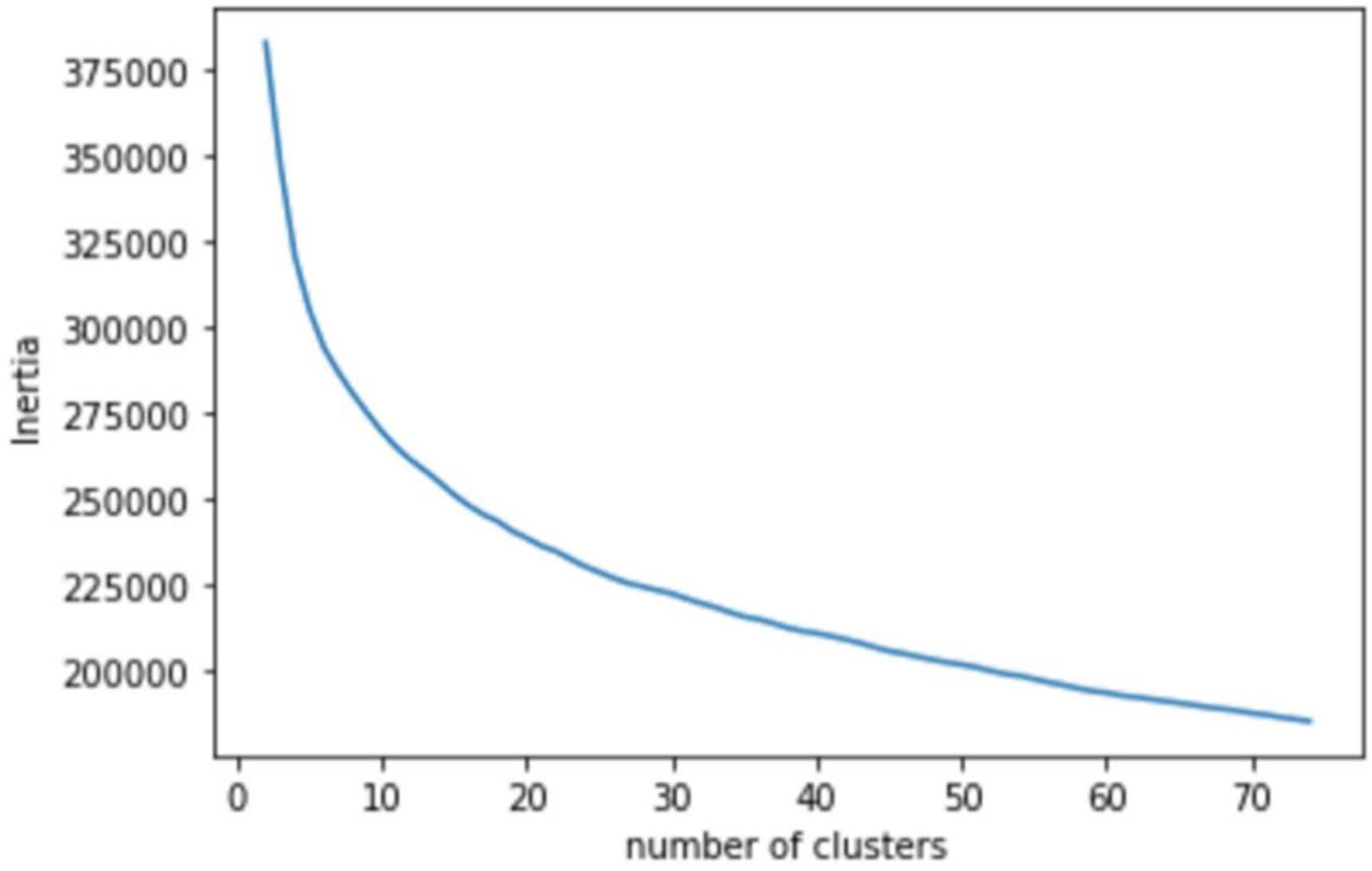

**Supplementary Figure S3**: Effect of changing the number of clusters on inertia quality metrics, which aggregates the distance from each datapoint to its centroid.

t-SNE

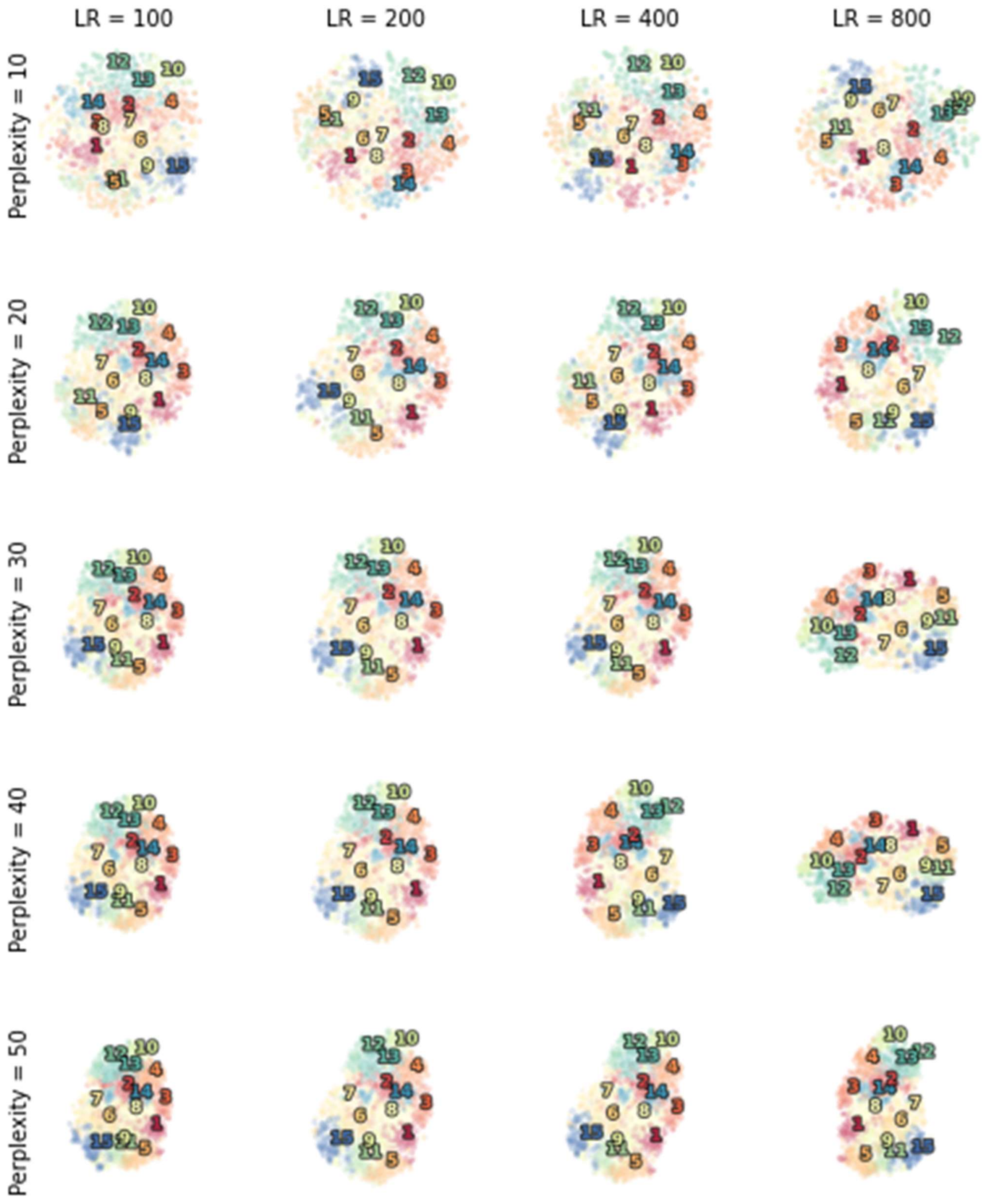

**Supplementary Figure S4**: Evaluation of t-SNE hyperparameters. To be able to confidently draw conclusions from the t-SNE plots, we also tuned the perplexity and learning rate (LR) hyperparameters. In the paper, we used perplexity 30 and LR 200.

T-distributed Stochastic Neighbor Embedding or t-SNE is a popular dimensional reduction method to visualize data. t-SNE takes high dimensional data and reduces it to a low dimensional graph that retains the relationship between the original datapoints. In our experiment, research abstracts were encoded in 200-dimension vectors and t-SNE was used to visualize these vectors in a two-dimensional space (**Figure 2**). This process requires many iterations and after each iteration, the t-SNE algorithm continuously aims to minimize data loss while trying to preserve the relationship between the datapoints. Here, we compare the effect of perplexity and learning rate (LR) on the shape of k=15 t-SNE plots to evaluate the appropriate parameter value. The LR hyperparameter in t-SNE is an essential parameter that affects the size of the changes made to stabilize the graph. Conversely, perplexity is a vital parameter that specifies the number of nearest neighbors each point in the dataset has (van der Maaten, 2008). A higher dimensional and larger dataset tends to require a larger perplexity value but typically it is between 5 to 50. Objective evaluation of t-SNE plots in **Supplementary Figure S4** shows that perplexity >= 30 produces stable partner at different LR. Some of these plots might be rotated or translated differently but the relationships between the datapoints are consistent. As a result, we decided to choose the default perplexity of 30 and LR of 200 for simplicity. Other t-SNE hyperparameters were set to default values as implemented in the scikit-learn library.

<u>Manual Validation</u>

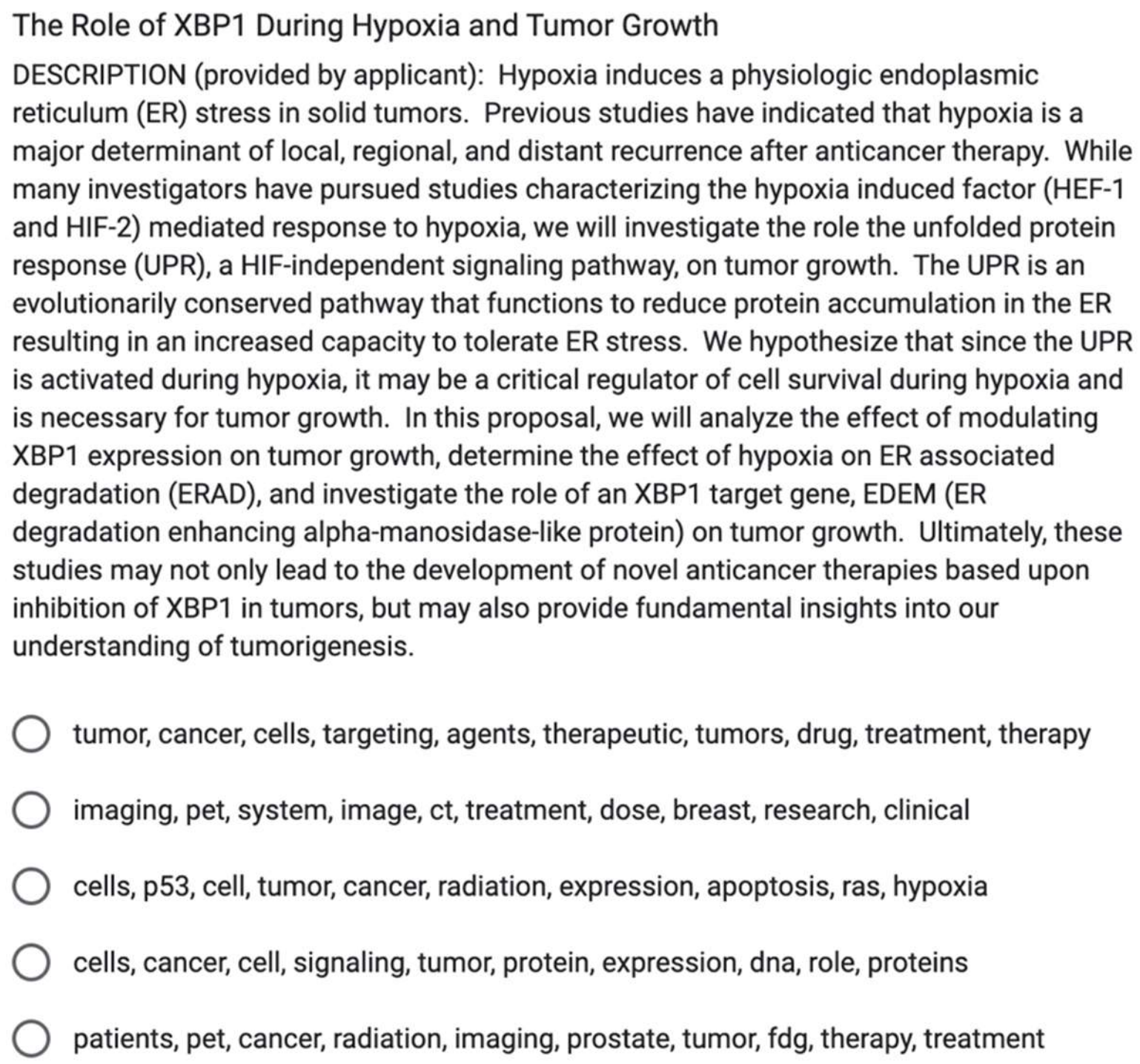
The Role of XBP1 During Hypoxia and Tumor Growth

DESCRIPTION (provided by applicant): Hypoxia induces a physiologic endoplasmic reticulum (ER) stress in solid tumors. Previous studies have indicated that hypoxia is a major determinant of local, regional, and distant recurrence after anticancer therapy. While many investigators have pursued studies characterizing the hypoxia induced factor (HEF-1 and HIF-2) mediated response to hypoxia, we will investigate the role the unfolded protein response (UPR), a HIF-independent signaling pathway, on tumor growth. The UPR is an evolutionarily conserved pathway that functions to reduce protein accumulation in the ER resulting in an increased capacity to tolerate ER stress. We hypothesize that since the UPR is activated during hypoxia, it may be a critical regulator of cell survival during hypoxia and is necessary for tumor growth. In this proposal, we will analyze the effect of modulating XBP1 expression on tumor growth, determine the effect of hypoxia on ER associated degradation (ERAD), and investigate the role of an XBP1 target gene, EDEM (ER degradation enhancing alpha-manosidase-like protein) on tumor growth. Ultimately, these studies may not only lead to the development of novel anticancer therapies based upon inhibition of XBP1 in tumors, but may also provide fundamental insights into our understanding of tumorigenesis.

- tumor, cancer, cells, targeting, agents, therapeutic, tumors, drug, treatment, therapy
- imaging, pet, system, image, ct, treatment, dose, breast, research, clinical
- cells, p53, cell, tumor, cancer, radiation, expression, apoptosis, ras, hypoxia
- cells, cancer, cell, signaling, tumor, protein, expression, dna, role, proteins
- patients, pet, cancer, radiation, imaging, prostate, tumor, fdg, therapy, treatment

**Supplementary Figure S5**: A sample prompt for manual validation of grant clustering. The title and text of the grant is shown. There are five multiple choice options representing 5 clusters. Each cluster was represented by the top 10 words by TF-IDF score.

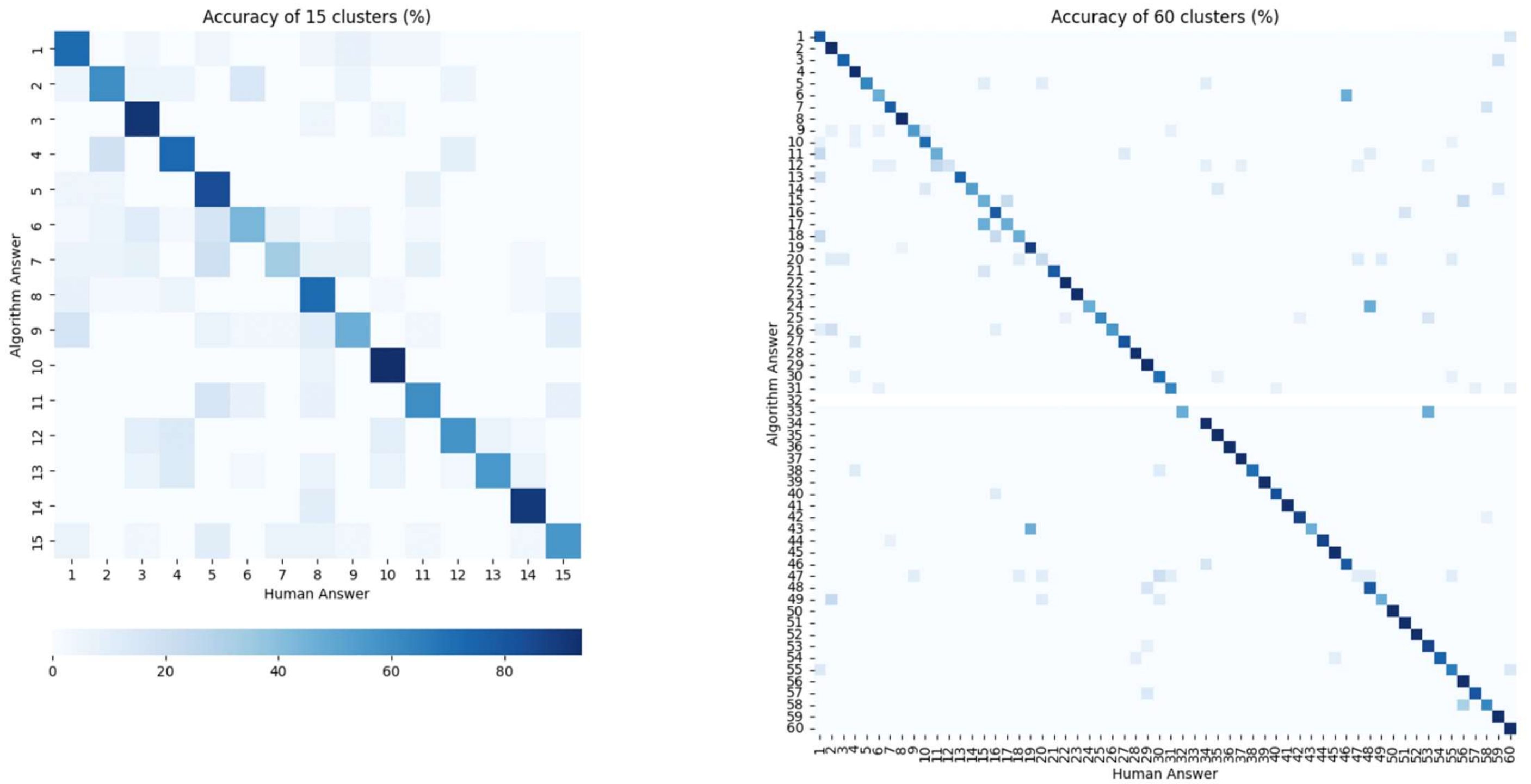

**Supplementary Figure S6**: Comparison of k=15 (left) and k=60 (right) clustering compared to human categorization for 400 manual reviews of grants (averaged over 4 human reviewers). Compared to k=60, k=15 had lower accuracy (62% vs. 77%) and Cohen's kappa agreement (53% vs 71%). This discrepancy was thought to be because at k=60, topics are more specific and are easier for our reviewers to validate. Cluster #32 (k=60) was not validated due to sampling error.

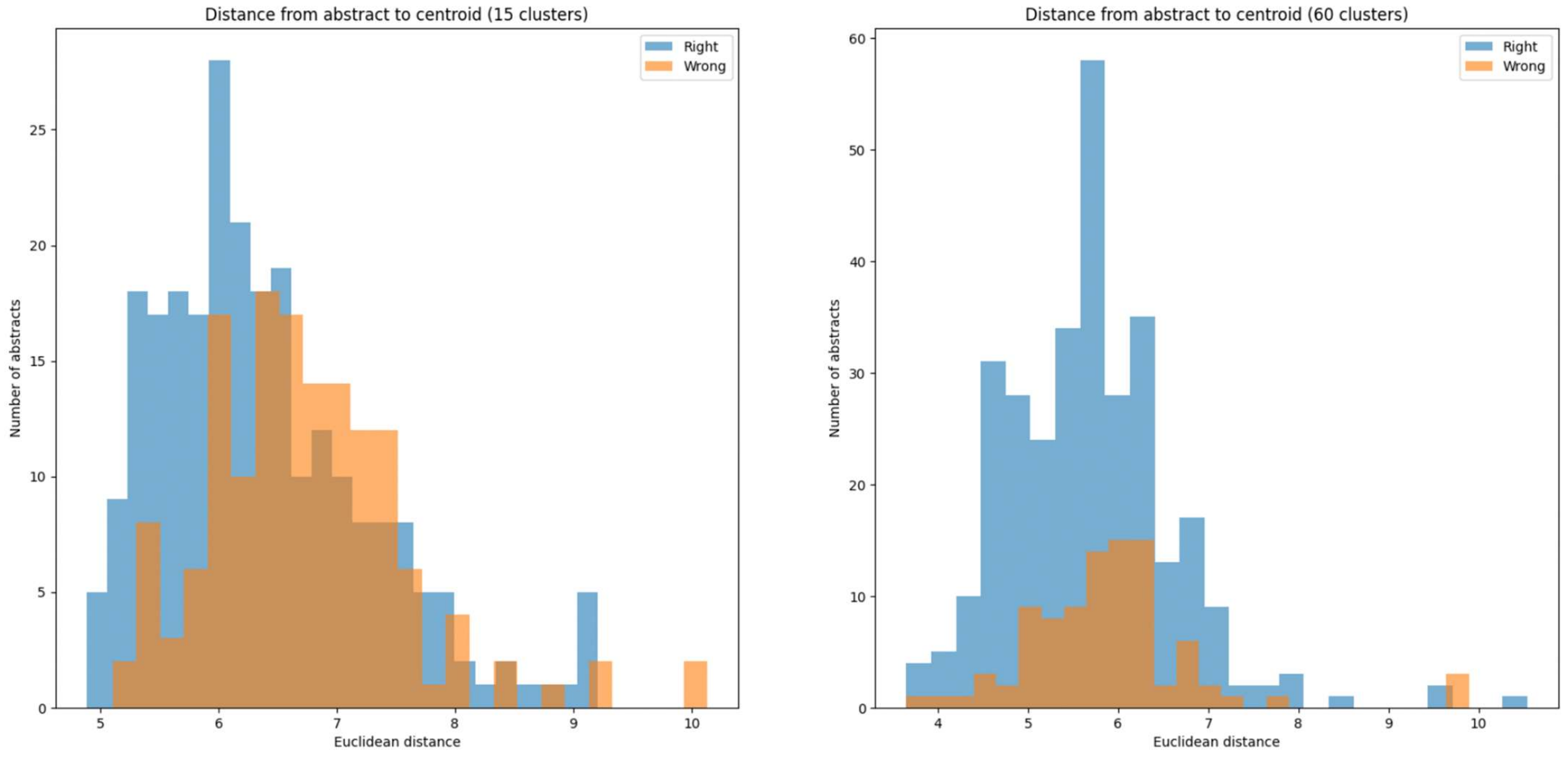

**Supplementary Figure S7**: Comparison of distance-to-centroid (or Euclidean distance) from abstract to centroid between correctly and wrongly selected answers when performing 400 manual validations for k=15 (left) and k=60 (right). For k=15, wrongly selected answers tend to have larger distance-to-centroids suggesting that at lower granularity (Kolmogorov-Smirnov test with p-value of 7.8e-6), abstracts that are less representative or further from the cluster centroid are more difficult to validate.

Breakdown of funding trends at high level of granularity

As the therapeutics-diagnostics and biology-physics axes split the t-SNE plot into four different quadrants, we also performed in depth analyses of these quadrants. Regarding the average growth rate per quadrant, there are inconsistent pattern between k=15 and k=60 topics, with diagnostics-physics quadrant and therapeutics-physics quadrant having the largest growth rate for k=15 (+$289,222 per year) and k=60 (+$102,881 per year). However, the growth rate is consistent between k=15 and k=60 when analyzing the biology-physics and diagnostics-therapeutics axes in isolation. Along the physics-biology axis, the grants on the physics (bottom) side of the t-SNE plots have higher growth rate for both k=15 (+$275,324 per year) and k=60 (+81,625 per year) compared to the biology grants. Similarly, along the diagnostics-therapeutics axis, the grants on the therapeutics (right) side of the t-SNE plots have higher growth rate for both k=15 (+$203,351 per year) and k=60 (+65,527 per year) compared to those on the diagnostics (left) side.

**Supplementary Table S2**: Average funding growth rate at different areas on the t-SNE plots.

| | Growth rate (k=15) (per year) | Growth rate (k=60) (per year) |
|---|---|---|
| Biology-Therapeutics (top right) | $154,498 | $32,779 |
| Biology-Diagnostics (top left) | $9,207 | $48,818 |
| Physics-Therapeutics (bottom right) | $259,076 | $102,881 |
| Physics-Diagnostics (bottom left) | $289,222 | $63,443 |
| Biology (top) | $87,668 | $40,157 |
| Physics (bottom) | $275,324 | $81,625 |
| Therapeutics (right) | $203,351 | $65,527 |
| Diagnostics (left) | $162,133 | $56,805 |